\documentclass[lettersize,journal]{IEEEtran}
\usepackage{amsmath,amsfonts}
\usepackage{algorithmic}
\usepackage{algorithm}
\usepackage{array}
\usepackage{multirow}
\usepackage{booktabs}
\usepackage[caption=false,font=normalsize,labelfont=sf,textfont=sf]{subfig}
\usepackage{textcomp}
\usepackage{stfloats}
\usepackage{url}
\usepackage{verbatim}
\usepackage{graphicx}
\usepackage{cite}
\usepackage{makecell}
\usepackage{bm}
\begin{document}

\title{Autonomous Manipulation of Hazardous Chemicals and Delicate Objects in a Self-Driving Laboratory: A Sliding Mode Approach}


\author{Shifa Sulaiman$^{1}$, Francesco Schetter$^{2}$, Tobias Jensen$^{1}$, Simon Bøgh$^{1}$, Fanny Ficuciello $^{2}$
\thanks{This work was supported by the Pioneer Center for Accelerating P2X Materials Discovery, CAPeX, DNRF grant number P3. $^{1}$Department of Electronic Systems, Aalborg University, Aalborg, Denmark. $^{2}$Department of Information Technology and Electrical Engineering, University of Naples, Naples, Italy.*ssajmech@gmail.com}
}



\maketitle

\begin{abstract}

Precise handling of chemical instruments and materials within a self-driving laboratory environment using robotic systems demands advanced and reliable control strategies. Sliding Mode Control (SMC) has emerged as a robust approach for managing uncertainties and disturbances in manipulator dynamics, providing superior control performance compared to traditional methods. This study implements a model-based SMC (MBSMC) utilizing a hyperbolic tangent function to regulate motions of a manipulator mounted on a mobile platform employed inside a self-driving chemical laboratory. Given the manipulator’s role in transporting fragile glass vessels filled with hazardous chemicals, the controller is specifically designed to minimize abrupt transitions and achieve gentle, accurate trajectory tracking. The proposed controller is bench marked against a non-model-based SMC (NMBSMC) and a Proportional-Integral-Derivative (PID) controller using a comprehensive set of joint and cartesian metrics. 
Compared to PID and NMBSMC, MBSMC achieved up to $10^3$ smoother motion and 90\% lower control effort, validating its robustness and precision for autonomous lab operations.
Experimental trials confirmed successful execution of tasks such as vessel grasping and window operation, which failed under PID control due to its limited ability to handle nonlinear dynamics and external disturbances, resulting in significant trajectory tracking errors. The results validated the controller’s effectiveness in achieving smooth, precise, and safe manipulator motions, supporting the advancement of intelligent mobile manipulators in autonomous laboratory environments.

\end{abstract}


\section{Introduction}
\IEEEPARstart{T}{he} automation of laboratory processes has gained significant momentum in recent years, driven by the need for enhanced reproducibility, efficiency, and safety in chemical experimentation \cite{pso}. Research laboratories across the globe are evolving into automated environments, with an integration of mobile manipulators \cite{mob1} in self-driving laboratories marking a major milestone in this transformation. Unlike traditional static robotic arms \cite{mob2}, mobile manipulators combine mobility and dexterity, allowing them to navigate autonomously within laboratory environments and execute complex tasks with minimal human supervision. Similar to mobile manipulators, humanoid robots are increasingly being studied as potential agents for self‑driving laboratories, where they can autonomously perform complex experimental tasks \cite{mob3, mob4}.
This shift towards robotic automation not only improves productivity but also minimizes errors and contamination risks associated with manual handling \cite{refnew1}.

In typical laboratory settings, many of the vessels used particularly brittle glass containers filled with reactive or hazardous chemicals introduce strict handling constraints. To prevent glass breakage and chemical spills, the manipulator's trajectory and motion profiles must be exceptionally smooth and stable. Conventional control strategies, such as Proportional-Integral-Derivative (PID) controllers, have historically been used to regulate robotic motion and manipulation. However, PID controllers struggle with disturbances, uncertainties in dynamic environments, and non-linear system behavior commonly found in chemistry labs. These limitations necessitate the adoption of more advanced control strategies, such as Sliding Mode Control (SMC), which is highly robust against system variations and disturbances. The ability of SMC to preserve stable manipulator behavior in the presence of fragile labware and sensitive chemical payloads makes it a highly suitable candidate for next-generation autonomous laboratory applications.
Major contributions of this work are as follows:
\begin{itemize}
    \item \textbf{Mobile Manipulator in Autonomous Chemical Lab:} Development of a mobile manipulator to employ inside a self-driving laboratory and addressing challenges in integrating such a robotic system along with a chemical experimentation workflow.
    
    \item \textbf{Hyperbolic Tangent-Based Model SMC:} A novel model-based SMC utilizing a hyperbolic tangent function is proposed to suppress chattering and ensure smooth manipulator trajectories, particularly critical for handling fragile chemical glassware.

    \item \textbf{Evaluation Using Higher-Order Motion Metrics:} Performance of the controller is analyzed using advanced smoothness metrics such as jerk and snap, in addition to velocity, acceleration, and standard control metrics.

    \item \textbf{Comprehensive Validation:} The proposed controller is validated both through simulation and experimental trials, showing consistent improvements in trajectory tracking and stability.

\end{itemize}

 In this study, a model-based SMC (MBSMC) is compared with a first-order non-model-based SMC (NMBSMC) and a traditional PID controller to evaluate their effectiveness in pick-and-place operations within a self-driving chemistry laboratory. Through simulation studies and experimental trials, this work assesses the ability of each control strategy to achieve stable and precise object manipulation in laboratory settings. Simulations provide insights into trajectory accuracy, control effort, and robustness against disturbances, while real-world experiments validate the manipulator's performance under laboratory conditions.  The comparative analysis demonstrated that the MBSMC outperformed both the PID and NMBSMC controllers in terms of precision and stability, making it a superior choice for self-driving lab automation. By integrating mobile manipulator with advanced control algorithms, this research contributes to the future development of fully autonomous laboratories. 

\section{Background}

Mobile manipulators, combining a robotic arm with a mobile platform, are employed in industrial automation, logistics, and healthcare fields \cite{ref2} . Their application in chemistry labs necessitates precise control mechanisms due to the delicate nature of the materials being handled. Several studies have focused on trajectory planning, collision avoidance, and adaptive control for such systems. Burger \textit{et al.} \cite{ref3} introduced a fully autonomous mobile robot, designed to accelerate materials discovery through experimentation. The system integrated a KUKA LBR IIWA robotic arm mounted on a non-holonomic KUKA KMP mobile platform, enabling it to autonomously navigate a laboratory, manipulate reagents, operate instruments, and perform complex chemical tasks. Using a Bayesian search algorithm, the robot conducted 688 photocatalysis experiments over eight days, optimizing hydrogen production from water. 
A mobile manipulator actively used in pharmaceutical and chemical laboratories to transport SBS-format plates and interface with formulation and analytical equipment was given in \cite{ref5}. It featured a robotic arm, a refrigerated storage, and 2D/3D vision systems. It performed intra-lab and inter-lab logistics, reducing manual transport, and enabling overnight autonomous operations. Despite its success, integration with Laboratory Information Management System (LIMS) and scheduling systems remains a challenge for full lab autonomy. 

Nguyen \textit{et al.} \cite{ref6} enhanced a TIAGO mobile manipulator by modeling non-geometric parameters such as gear backlash, encoder misalignment, and base suspension dynamics. Using a stereophotogrammetric system, they improved end-effector pose accuracy by 60\%, demonstrating the importance of accounting for mechanical imperfections in lab-grade mobile robots. 
The National Institute of Standards and Technology (NIST) developed a Configurable Mobile Manipulator Apparatus \cite{ref7} to measure continuous performance in large-scale lab environments. It integrated coordinate registration, Kalman filters, and optical tracking to assess pose repeatability and accuracy. This work addresses the lack of performance benchmarks for mobile manipulators in labs, especially for complex workpieces and multi-zone operation. The Enabled Robotics Platform \cite{ref8} combined a MiR200 mobile base with a UR5 manipulator, equipped with an OnRobot RG6 gripper and a vision system. It was used in collaborative lab environments for tasks like sample handling, equipment calibration, and automated logistics. Its modularity allowed for easy adaptation, but the study noted limitations in whole-body motion coordination and dynamic obstacle avoidance. Jiang \textit{et al.} \cite{ref11_new}  presented a dual-arm arm employed for solid dispensing tasks. The proposed system was able to transfer solids in milligram level precisions.

SMC methods have been successfully applied to robotic manipulators to ensure robust trajectory tracking and disturbance rejection, even in the presence of modeling uncertainties and nonlinear dynamics. 
A neuro-adaptive SMC was applied to a manipulator for sample handling tasks in \cite{ref9}. Using generalized artificial neural networks, the controller achieved precision with reduced steady-state error. Results indicated a 16.4\% reduction in tracking error and a 45\% faster response, reinforcing the controller’s feasibility for dynamic lab automation. The system lacked payload adaptation under real-time conditions a gap in resilience.
Babaei \textit{et al.} \cite{ref10} employed a fast terminal SMC combined with a nonlinear disturbance observer for trajectory tracking in robotic manipulators handling chemical samples. Implemented on a 3-Degrees of Freedom (DOF) arm, the system demonstrated a 55\% increase in tracking accuracy and a 45\% decrease in response time. While robust to noise and dynamics, scalability to more complex manipulators and elimination of residual chattering remain unresolved. Li and Song \cite{ref11} developed a discrete adaptive terminal SMC for a two-joint manipulator. The controller achieved tracking accuracy within 0.004 radians, with control torque variation minimized to $5.85 \times 10^{3}~Nm$. 

Park \textit{et al.} \cite{ref12} integrated a super-twisting SMC with adaptive neural network compensators for mobile manipulators executing material transport across lab stations. The control strategy excelled in maintaining trajectory accuracy despite wheel-ground slippage, validating its practicality for real lab terrains. However, issues of network generalization across different floor textures and retraining overhead persist. Yun and Lee \cite{ref13} explored Convolutional Neural Network (CNN)-enhanced SMC for dynamic uncertainty compensation in wall-climbing robotic manipulators used in surface inspection. Though originally developed for welding, the technique was adapted to lab inspection scenarios requiring meticulous scanning. The controller reduced chattering and improved stability, but data-intensive CNN training requirements limit its scalability in settings with sparse datasets.

Across previous research, several notable limitations have emerged in the SMC strategies applied to laboratory manipulators. These include the persistent challenge of high-frequency chattering, the lack of real-time adaptability to dynamic payload changes, limited scalability beyond low DOF systems, and heavy reliance on simulation data without sufficient experimental validation in live laboratory settings. 
In this paper, we aim to directly address these gaps by developing an SMC framework that integrates lightweight adaptive compensation for dynamic uncertainties, ensures smooth control transitions to suppress chattering, and is experimentally deployed on a 6-DOF mobile manipulator system. Our methodology emphasizes performance evaluation under realistic lab conditions, prioritizing precision, robustness, and controller efficiency to bridge the gap between theoretical modeling and hands-on automation.

\section{Methodology}
\begin{figure}[htbp]
    \centering    \includegraphics[width=0.4\textwidth]{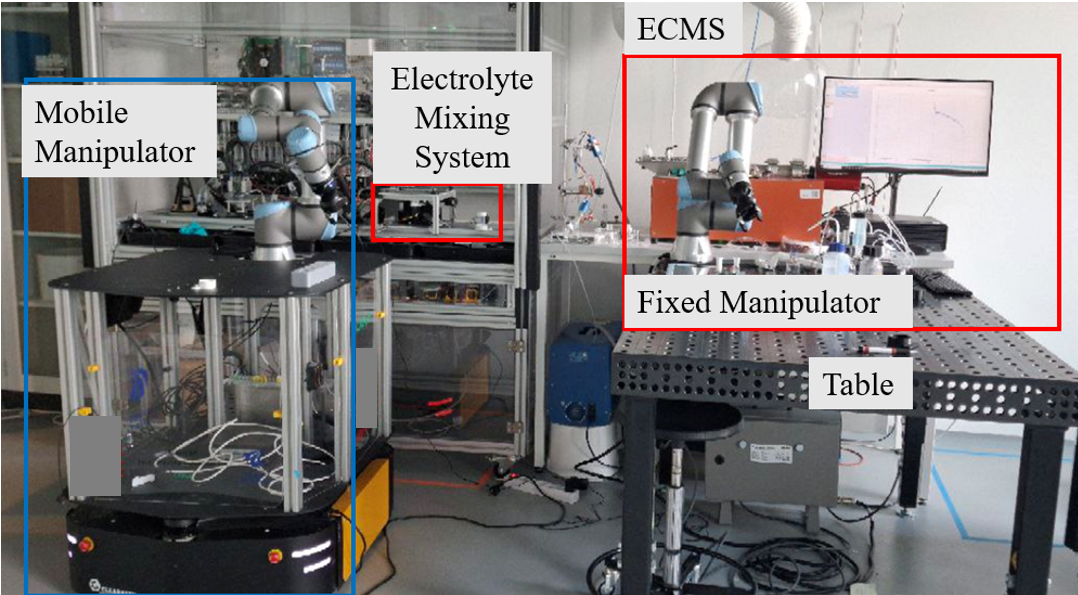}
    \caption{System overview and experimental setup showing mobile manipulator, electrolyte mixing system, fixed manipulator, and ECMS machine}
    \label{lab}
\end{figure}

\begin{figure}[htbp]
    \centering   \includegraphics[width=0.35\textwidth]{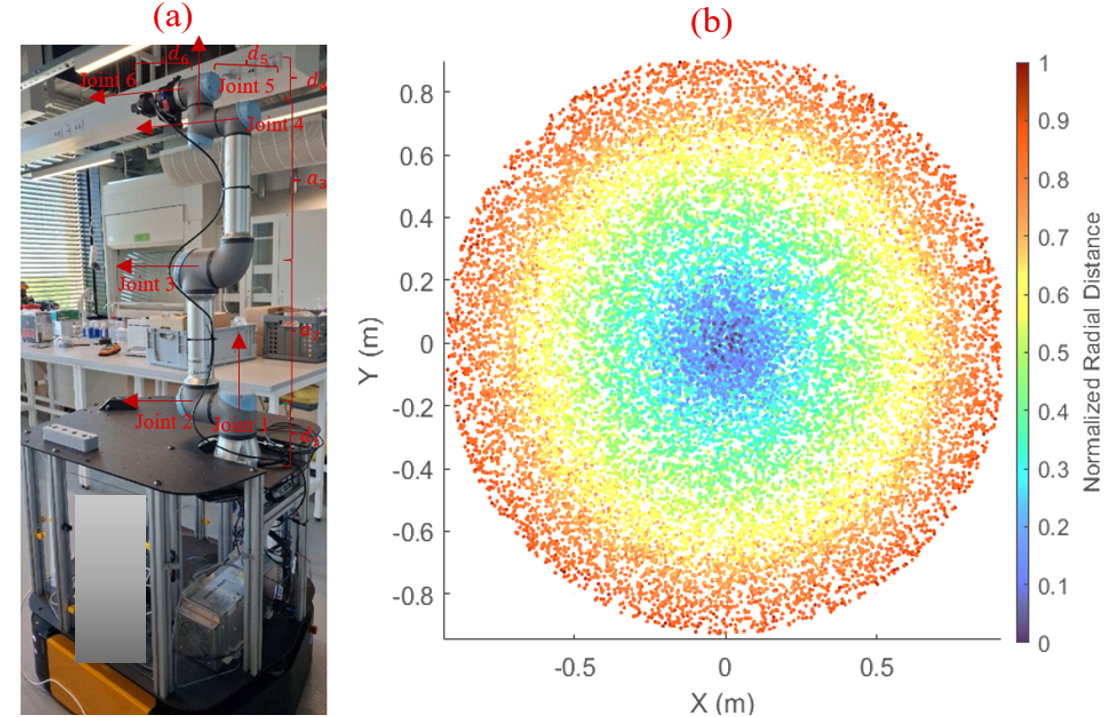}
    \caption{Mobile manipulator (a)Frames of manipulator (b)Workspace of manipulator}
    \label{mobile manipulator}
\end{figure}

\begin{figure}[htbp]
    \centering  \includegraphics[width=0.5\textwidth]{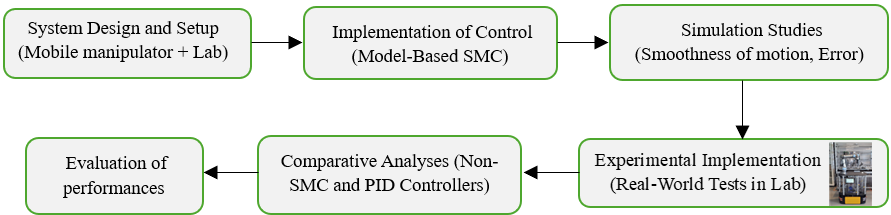}
    \caption{Methodology and control framework}
    \label{Methodology}
\end{figure}

The methodology adopted in this paper follows a structured approach to develop, implement, and evaluate a mobile manipulator robot equipped with SMC within a self-driving chemistry laboratory. The process begins with designing a mobile manipulator capable of autonomously navigating and performing pick-and-place tasks in a dynamic laboratory environment as shown in Fig. \ref{lab}. The laboratory environment consisted of a mobile manipulator, a fixed-base manipulator, an electrolyte mixing system, and an   Electrochemical Mass Spectrometry (ECMS) machine. The mobile manipulator system comprised of a Ridgeback omnidirectional platform integrated with a Universal Robot robotic arm (UR5e), equipped with a Robotiq Hand-E adaptive gripper as shown in Fig. \ref{mobile manipulator}(a). 
The Ridgeback base provides smooth and holonomic navigation with a high payload capacity, enabling precise maneuverability within cluttered lab environments. Mounted atop is the UR5e arm, a 6-DOF collaborative manipulator renowned for its accuracy and integrated force-torque sensing, making it well-suited for delicate interactions. The top view of the Cartesian workspace of the manipulator (colour grading with respect to radial distance) is shown in Fig. \ref{mobile manipulator}(b). The UR5e manipulator has an estimated workspace volume of approximately $2.27~\text{m}^3$, shaped by its $850~\text{mm}$ reach and constrained by joint limits and tool configuration. At the end-effector, the Robotiq Hand-e gripper offers versatile gripping capabilities, accommodating a wide range of labware, from small vials to larger containers. 
Enhanced with onboard sensors and software package like ROS MoveIt, the system achieves robust localization, real-time motion planning, and obstacle avoidance, forming a reliable platform for intelligent lab automation. The system is equipped with 2 LIDAR sensors and three real-sense cameras. 
The steps adopted in the methodology is shown in Fig. \ref{Methodology}. 

An MBSMC was developed incorporating detailed system dynamics to control motions of the manipulator handling chemical laboratory tasks. 
To validate the effectiveness of these control strategies, simulations are conducted using  physics-based robot simulation platform like Gazebo. The simulations allow for an initial assessment of how each controller responds to disturbances, trajectory tracking requirements, and accuracy in object manipulation. 
Following simulation validation, experimental trials are carried out in a chemistry laboratory environment. The mobile manipulator is tested on various pick-and-place tasks, including handling chemical containers, moving laboratory instruments, and interacting with automated electrolyte mixing systems. The real-world experiments enabled a direct comparison between the three controllers, with data collected through trajectory tracking systems, force sensors, and visual feedback mechanisms. The experimental setup ensured that results reflected practical laboratory conditions, highlighting key differences in controller performance when exposed to environmental uncertainties.
Finally, a comparative analysis of the the proposed control strategy with NMBSMC and PID controllers is performed to determine the performance of controllers. 

\section{Sliding Mode Control Strategy for UR5e }
\begin{figure}[htbp]
    \centering   \includegraphics[width=0.35\textwidth]{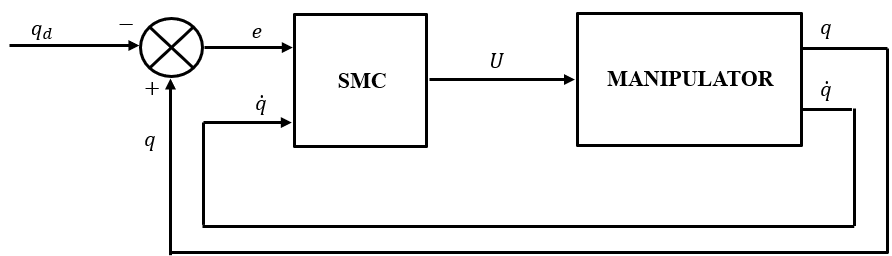}
    \caption{SMC controller framework for UR5e manipulator}
    \label{hybrid}
\end{figure}

Modern robotic manipulators are increasingly expected to operate reliably in environments laden with uncertainties, dynamic disturbances, and nonlinear behaviors. The UR5e is  known for its flexibility and ease of programming, serves as a suitable platform for exploring advanced control strategies aimed at enhancing precision and robustness in joint-space control. 
For UR5e joints, implementing an SMC framework provides precise trajectory tracking and strong disturbance rejection. 
An SMC strategy was developed to enhance the precision and robustness of the UR5e manipulator under dynamic operating conditions. 
The dynamic model of UR5e with joint positions ($q$), joint velocities ($\dot{q}$), and joint accelerations ($\ddot{q}$) is given in equation \eqref{dynmc}.
\begin{equation}
{M(q)}\ddot{q} + C(q,\dot{q})\dot{q} + G(q) = \tau
\label{dynmc}
\end{equation}
where $M(q)$, $C(q,\dot{q})$, $G(q)$, and $\tau$ are the inertia matrix, Coriolis-centrifugal term matrix, gravity matrix, and input joint torque vector respectively.
The proposed control architecture (illustrated in Fig.~\ref{hybrid}) combines a reduced-order dynamic model with SMC to maintain accurate and stable motion across tasks.
SMC block receives angular velocity ($\dot q)$ and error ($e$ = output joint angles $(q)$ - desired joint angles $(q_d)$) as the inputs. 
The SMC block generates joint actuation torque ($\tau = U$), which drives the UR5e joints. 
For state-space representation, we define:
\begin{align}
    x_1 &= q \label{eq_1} \\
    x_2 &= \dot{q} \label{eq_2} \\
    \dot{x}_2 &= \ddot{q} \label{eq_3}
\end{align}
 Starting from equation \ref{dynmc}, by denoting $\tau$ as $U$,the linear state-space form becomes:
\begin{align}
    \dot{x}_1 &= x_2 \notag \\
    \dot{x}_2 &= M^{-1}(U-Cx_2-G)
    \label{ssf}
\end{align}
Lets define the system's state vector as follows:
\begin{equation}
    \bm{x} =
        \begin{bmatrix}
        x_1 \\
        x_2
    \end{bmatrix}
    =
    \begin{bmatrix}
        q \\
        \dot{q}
    \end{bmatrix}
    \label{st}
\end{equation}
The nonlinear dynamic form is given in equation \eqref{eq:13}.
\begin{equation}
    \dot{\bm{x}} = \bm{f(x)} + \bm{g(x)} U
    \label{eq:13}
\end{equation}
where $f(x)$ is the vector field and $g(x)$ is the function which maps $U$ to the force field of the system. Both are given in equation \eqref{eq:14}.
\begin{equation}
    \bm{f(x)} =
    \begin{bmatrix}
        x_2 \\
        -M^{-1}(Cx_2+G)
    \end{bmatrix}, \quad
    \bm{g(x)} =
    \begin{bmatrix}
        0 \\
        M^{-1}
    \end{bmatrix}
    \label{eq:14}
\end{equation}
The control input $U$, which is the summation of equivalent control ($U_{equivalent}$) and switching control ($U_{switching}$), is formulated as follows:
\begin{equation}
    U = \left(U_{\text{equivalent}} + U_{\text{switching}}\right)
    \label{eq:16}
\end{equation}
where $U_{equivalent}$ is given in equation \eqref{eq:17}.
\begin{equation}
    U_{\text{equivalent}} =- \frac{L_f(\sigma(x))}{L_g(\sigma(x))}
    \label{eq:17}
\end{equation}
Sliding surface function ($\sigma(x)$) is given in equation \eqref{eq:18}.

\begin{equation}
    \sigma(x) = P_1 e(t) + P_2 x_2
    \label{eq:18}
\end{equation}
where $P_1$ and $P_2$ are controller tuning parameters used to regulate convergence and robustness. The sliding manifold, $S$ is defined as in equation \eqref{eq:19}.
\begin{equation}
    S = \left\{ \bm{x} \in \mathbb{R}^n : \sigma(x) = 0 \right\}
    \label{eq:19}
\end{equation}
To minimize chattering, the switching control component is defined as follows:
\begin{equation}
    U_{\text{switching}} = -\frac{P_3}{L_g(\sigma(x))} \tanh(\sigma(x))
    \label{eq:20}
\end{equation}
where $P_3$ fine-tunes the system’s convergence rate. The system's vector field can be decomposed into \( f^+(x) \) and \( f^-(x) \), as defined in equation~\eqref{eq_f+-}.

\begin{equation}
f(x)=
    \begin{cases}
        f^+(x)=f(x)+g(x)U^+\\
        f^-(x)=f(x)+g(x)U^-
    \end{cases}
    \label{eq_f+-}
\end{equation}
Finally, the control input is applied based on the sign of the sliding surface given in equation \eqref{eq:21}.
\begin{equation}
    U =
    \begin{cases}
        U^+, & \sigma(x) > 0 \\
        U^-, & \sigma(x) < 0
    \end{cases}
    \label{eq:21}
\end{equation}
The control input \( U \) should be designed such that the vector fields \( f^+(x) \) and \( f^-(x) \) are not only attracted toward the sliding surface \( S \), but also constrained to slide along it. Therefore, to achieve sliding mode control, it is essential to define \( U \) in a way that satisfies the following conditions:
\begin{itemize}
    \item Vector field of the system must be tangent to the sliding surface.
    \item The sliding surface should exert attractive force.
\end{itemize}
In order to satify the first requirement, we have to consider a 'sliding vector field', $f_s(x)$, defined using Filippov Convexification form (equation \ref{eq_fs})
\begin{equation}
    f_s(x)=\alpha f^+(x)+(1-\alpha)f^-(x),~\alpha \in [0,1] 
    \label{eq_fs}
\end{equation}
If equation~\eqref{eq_fs} is updated based on equation~\eqref{eq_f+-}, we obtain:

\begin{equation}
    f_s(x)=f(x)+g(x)U_{equivalent}
\end{equation}
where
\begin{equation}
    U_{equivalent}=\alpha U^+ +(1-\alpha)U^-, ~\alpha \in [0,1] 
\end{equation}
Hence, $U_{eq}$ should be chosen in such a way  that $f_s(x)$ is tangent every time to $S$ (Utkin's equivalent control)\cite{utkin2004sliding}. In order to ensure this, consider the Lie derivative of $f_s(x)$ with respect to $\sigma(x)$ that leads to equation \eqref{eq:lie}.
\begin{equation}
    L_{f_s}(\sigma)=L_f(\sigma)+L_g(\sigma)U_{equivalent}
    \label{eq:lie}
\end{equation}
Since the Lie derivative \( L_f(\sigma) = \nabla \sigma \cdot f(x) \) describes the direction of the vector field \( f(x) \) relative to the surface, where \( \nabla \sigma \) is the normal vector at each point. Equation~\eqref{eq:lie} must be zero to satisfy the tangency condition, and Equation~\eqref{eq:17} can be used to find an optimized value of \( U_{\text{equivalent}} \), provided that the transversality condition \( L_g(\sigma) \neq 0 \) is met, as failure to verify this condition may render the influence of the vector field \( g(x) \) ineffective, as a zero Lie derivative \( L_g(o(x)) = 0 \) implies that the control input \( U \) has no directional effect on the system's motion along the sliding surface. In such cases, the controller cannot enforce sliding mode behavior, compromising stability and tracking performance. For satisfying the second condition we can consider the Lyapunov function given in equation \eqref{eq:lya}
\begin{equation}
    V=\frac{1}{2}\sigma(x)^2
    \label{eq:lya}
\end{equation}
Differentiation of equation \eqref{eq:lya} leads to following equation:
\begin{equation}
    \dot V= \sigma(x) \dot \sigma(x) = \sigma(x) \nabla \sigma \dot x
\end{equation}
Substituing in equation (\ref{eq:13}) we get:
\begin{equation}
    \dot V = \sigma(x) \nabla \sigma (f(x)+g(x)U)= \sigma(x)(L_f(\sigma)+L_g(\sigma)U) 
\end{equation}
In order to guarantee the actions of the sliding surface, $\dot V < 0$ and this leads to:
\begin{equation}
    \begin{cases}
        L_f(\sigma)+L_g(\sigma)U<0, & \sigma(x)>0\\
        L_f(\sigma)+L_g(\sigma)U>0, & \sigma(x)<0
    \end{cases}
\end{equation}
If the control action \( U \) is selected according to equation~\eqref{eq:16}, where \( U_{\text{equivalent}} \) is obtained from equation~\eqref{eq:17} and \( U_{\text{switching}} \) is defined by equation~\eqref{eq:20}, then the derivative of the Lyapunov function satisfies \( \dot{V} = -P_3 |\sigma(x)| \), thereby fulfilling the second condition. This controller design also ensures that system trajectories reach the sliding surface in finite time. Specifically, we observe the following conditions:
\begin{gather}
    \dot V= \sigma(x) \dot \sigma(x) = -P_3 |\sigma(x)|\\
    \dot \sigma(x)= -P_3 \sigma(x) \to \sigma(x(t))= -P_3 t + \sigma(x_0)
    \label{eq_sigmat}
\end{gather}
By evaluating the equation \ref{eq_sigmat} we can determine following equation:
\begin{equation}
t=
    \begin{cases}
        t^+=\frac{\sigma(x_0)}{P_3} & \sigma(x_0)>0\\
        t^-=-\frac{\sigma(x_0)}{P_3} & \sigma(x_0)<0\\
    \end{cases}
\end{equation}
where $x_0$ is the initial position and the control action $U$ is chosen as follows:
\begin{equation}
    U=-\frac{P_1}{P_2}M+Cx_2+G-\frac{M}{P_2}P_3Tanh(\sigma)
\end{equation}
This SMC framework ensures precise and resilient control of the UR5e, optimizing joint tracking under variable conditions.
For the non model-based version of SMC, the control action $U$ is chosen as proportional action respect to position error and a switching term for chattering purpose as given in equation \eqref{nsmc}
\begin{equation}
    U = P_1e(t) + P_3Tanh(\sigma)
    \label{nsmc}
\end{equation}
In PID control strategy , the control action $U$ is chosen as follows:
\begin{equation}
U = K_Pe(t)+K_I\int_o^t e(t)dt~+K_D\dot{e}(t)
\label{pid}
\end{equation}

\section{Results and Discussion}
\begin{figure}[htbp]
    \centering   \includegraphics[width=0.4\textwidth]{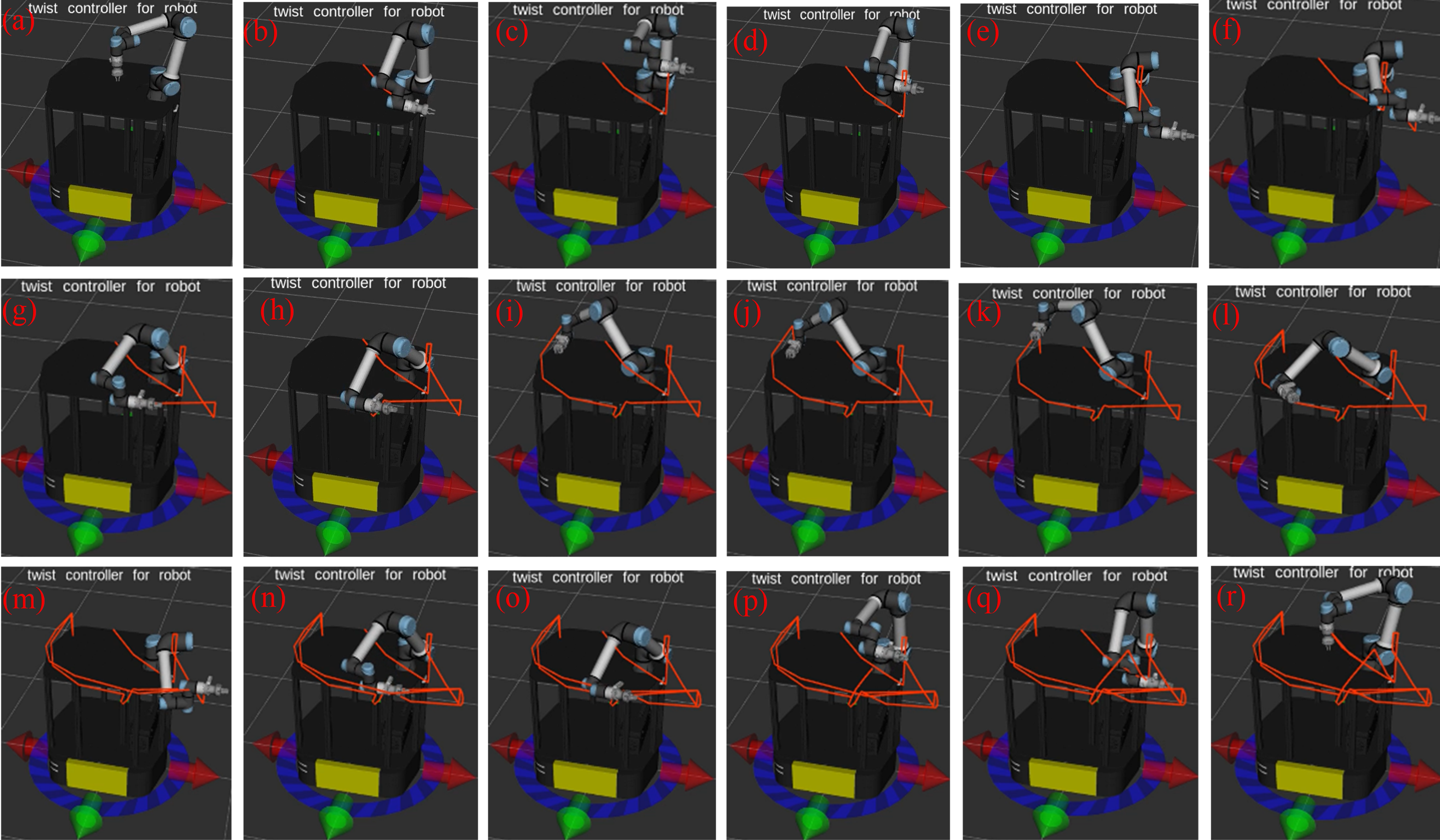}
    \caption{Motion of manipulator (base is stationary)}
    \label{sim}
\end{figure}

\begin{figure}[htbp]
    \centering   \includegraphics[width=0.4\textwidth]{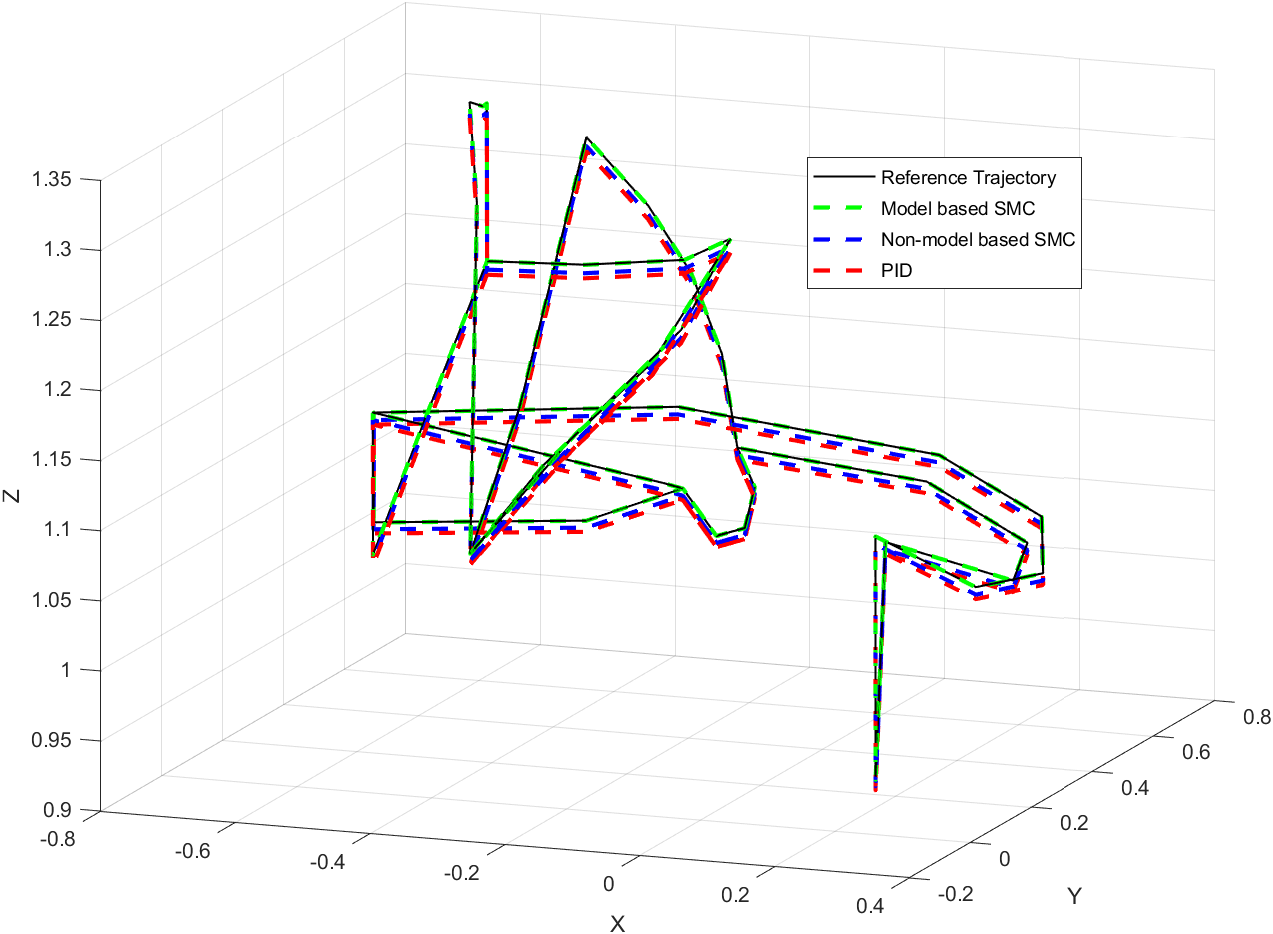}
    \caption{Comparison of Trajectories obtained using MBSMC, NMBSMC, and PID during simulation}
    \label{Traj_sim}
\end{figure}

\begin{figure}[htbp]
    \centering   \includegraphics[width=0.4\textwidth]{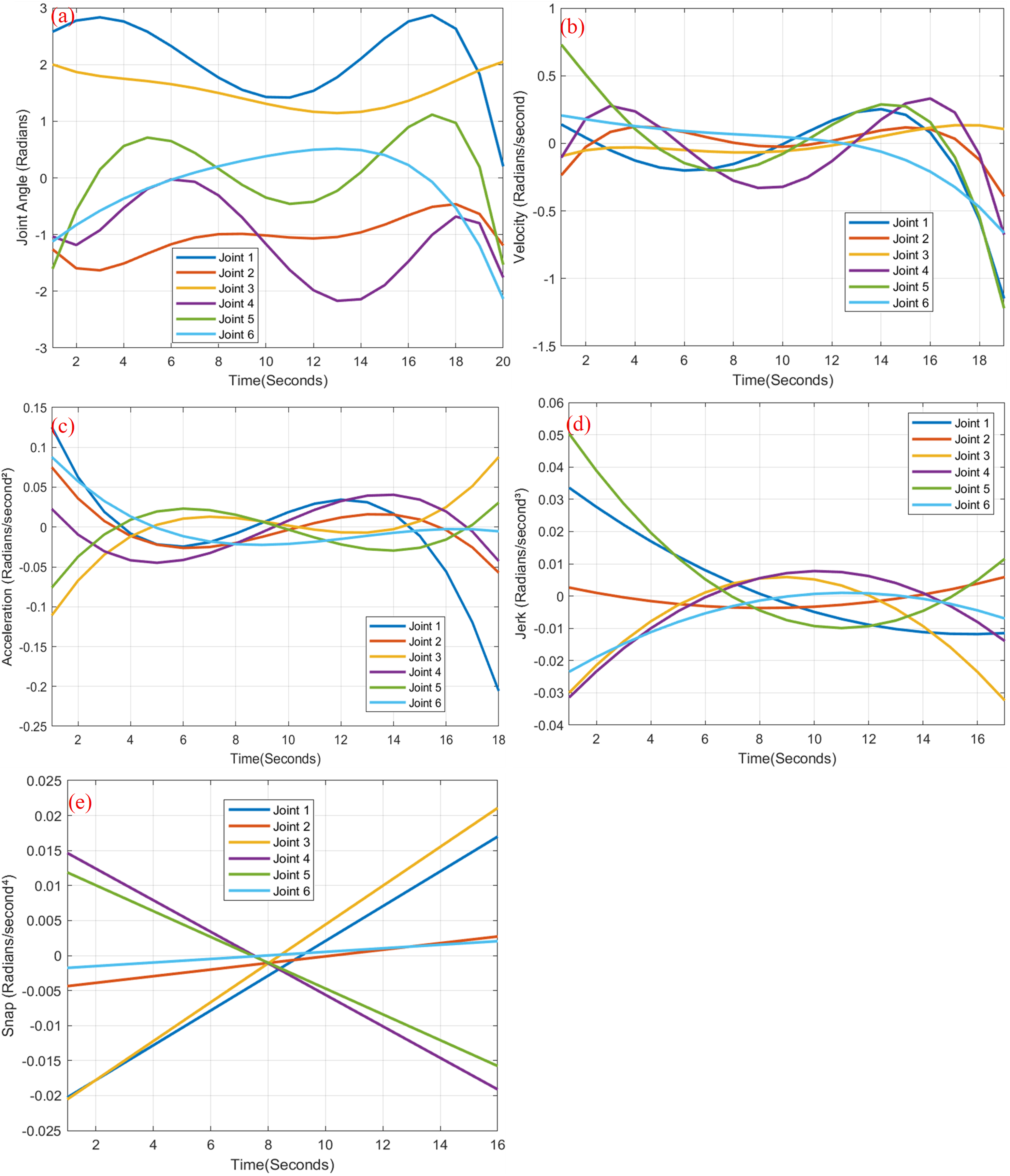}
    \caption{Simulation results using MBSMC (a)Joint angles (b)Velocity (c)Acceleration (d)Jerk (e)Snap}
    \label{sim_res15}
\end{figure}

Simulation studies and experimental validations were carried out to compare the performances of MBSMC against NMBSMC and PID controllers. These controllers were selected to represent widely adopted strategies in robotics. PID for its simplicity and ease of implementation, and NMBSMC for its robustness without relying on system dynamics.
Each controller was assessed based on its ability to produce smooth and accurate joint-space and Cartesian-space motions while responding to disturbances and varying input conditions. Evaluation metrics included motion smoothness quantified via velocity, acceleration, jerk, and snap profiles, trajectory tracking performance measured through Root Mean Square Error (RMSE).  Velocity continuity, acceleration profiles, jerk, and snap values were calculated by determining the maximum difference between successive joint values. The manipulator was assumed to be traversing a trajectory as shown in Fig. \ref{sim}. 
RRT* algorithm was used for determining the trajectory and the inverse solutions were obtained using Damped Least Square approach. By effectively handling the nonlinearities and uncertainties inherent in the manipulator's dynamic behavior, the SMC enabled smoother and more efficient motion, enhancing overall system performance. The tuning parameters ($P_1$, $P_2$, and $P_3$), initially set to 1, were instrumental in balancing responsiveness and stability. To further minimize output error and suppress chattering, these parameters were optimized using Particle Swarm Optimization (PSO) \cite{pso1}, resulting in final values of 100, 1, and 60 for $P_1$, $P_2$, and $P_3$, respectively. The trajectories obtained using the application of controllers in comparison to the reference trajectory are shown in Fig. \ref{Traj_sim}.
Joint angle, velocity, acceleration, jerk and snap values during the motion of manipulator obtained using proposed MBSMC are shown in Figs. \ref{sim_res15} (a) - (e). 
The performance metrics including RMSE in joint motions and Cartesian coordinates of the desired trajectory, steady-state errors, and the smoothness of joint motions characterized by velocity continuity, acceleration profile, jerk, and snap achieved using the model-based SMC and two alternative controllers are summarized in Table~\ref{tab:controller_comparison_reduced}.
\newcolumntype{B}{!{\vrule width 1.5pt}}  
\begin{table*}[h!]
\centering
\caption{COMPARISON OF SIMULATION RESULTS}
\label{tab:controller_comparison_reduced}
\resizebox{0.95\textwidth}{!}{%
\begin{tabular}{|l|c|c|c|c|c|c B c|c|c|c|c|c B c|c|c|c|c|c|}
\hline
\multirow{2}{*}{\textbf{Metric}} 
 & \multicolumn{6}{c}{\textbf{MBSMC} } \setlength{\arrayrulewidth}{1.5pt}\vline
 & \multicolumn{6}{c}{\textbf{NMBSMC} } \setlength{\arrayrulewidth}{1.5pt}\vline
 & \multicolumn{6}{c}{\textbf{PID} } \setlength{\arrayrulewidth}{1.5pt}\vline \\
 \cline{2-19}
& $\theta_1$ & $\theta_2$ & $\theta_3$ & $\theta_4$ & $\theta_5$ & $\theta_6$ 
& $\theta_1$ & $\theta_2$ & $\theta_3$ & $\theta_4$ & $\theta_5$ & $\theta_6$
& $\theta_1$ & $\theta_2$ & $\theta_3$ & $\theta_4$ & $\theta_5$ & $\theta_6$ \\
\hline
RMSE of joint motions ($rad$) & $2.8\times 10^{-4}$ & $6.0\times 10^{-4}$ & $2.1\times 10^{-4}$ & $4.1\times 10^{-4}$ & $0.7\times 10^{-4}$ & $2.0\times 10^{-4}$ & $2.1\times 10^{-3}$ & $1.6\times 10^{-3}$ & $3.2\times 10^{-4}$ & $3.4\times 10^{-3}$ & $2.1\times 10^{-4}$ & $1.2\times 10^{-4}$ & $1.3\times 10^{-1}$ & $5.2\times 10^{-1}$ & $3.3\times 10^{-2}$ & $7.1\times 10^{-2}$ & $1.0\times 10^{-3}$ & $4.2\times 10^{-3}$ \\
Velocity Continuity ($rad/s$)   & $2.1\times 10^{-3}$ & $2.9\times 10^{-3}$ & $3.0\times 10^{-3}$ & $2.0\times 10^{-3}$ & $3.1\times 10^{-4}$ & $4.0\times 10^{-4}$ & $4.2\times 10^{-3}$ & $3.0\times 10^{-3}$ & $1.1\times 10^{-2}$ & $2.5\times 10^{-3}$ & $1.6\times 10^{-3}$ & $3.2\times 10^{-2}$ & $4.1\times 10^{-2}$ & $5.1\times 10^{-1}$ & $3.3\times 10^{-2}$ & $1.0\times 10^{-2}$ & $3.2\times 10^{-3}$ & $3.8\times 10^{-2}$ \\
Acceleration Profile ($rad/s^2$)  & $4.1\times 10^{-3}$ & $5.0\times 10^{-3}$ & $5.3\times 10^{-3}$ & $3.8\times 10^{-3}$ & $2.9\times 10^{-4}$ & $3.7\times 10^{-4}$ & $5.4\times 10^{-3}$ & $1.3\times 10^{-2}$ & $4.2\times 10^{-3}$ & $1.9\times 10^{-2}$ & $3.8\times 10^{-3}$ & $2.9\times 10^{-2}$ & $1.4\times 10^{-1}$ & $4.9\times 10^{-2}$ & $1.0\times 10^{-2}$ & $1.3\times 10^{-1}$ & $0.3\times 10^{-1}$ & $5.6\times 10^{-2}$ \\
Jerk Profile ($rad/s^3$)          & $2.0\times 10^{-4}$ & $4.0\times 10^{-4}$ & $1.6\times 10^{-3}$ & $1.9\times 10^{-3}$ & $1.4\times 10^{-4}$ & $2.0\times 10^{-4}$ & $6.0\times 10^{-3}$ & $9.0\times 10^{-3}$ & $2.1\times 10^{-3}$ & $4.3\times 10^{-3}$ & $1.6\times 10^{-2}$ & $1.9\times 10^{-3}$ & $1.5\times 10^{-1}$ & $2.0\times 10^{-2}$ & $6.0\times 10^{-3}$ & $9.0\times 10^{-3}$ & $1.0\times 10^{-1}$ & $3.6\times 10^{-3}$  \\
Snap Profile  ($rad/s^4$)         & $0.9\times 10^{-4}$ & $0.7\times 10^{-4}$ & $2.8\times 10^{-4}$ & $1.3\times 10^{-4}$ & $0.5\times 10^{-4}$ & $0.3\times 10^{-4}$ & $3.0\times 10^{-3}$ & $1.3\times 10^{-2}$ & $2.2\times 10^{-3}$ & $1.2\times 10^{-2}$ & $3.2\times 10^{-2}$ & $1.6\times 10^{-3}$ & $2.0\times 10^{-2}$ & $1.8\times 10^{-1}$ & $3.3\times 10^{-2}$ & $3.0\times 10^{-2}$ & $2.6\times 10^{-1}$ & $5.6\times 10^{-2}$ \\
Steady state error ($rad$)    & $0.3\times 10^{-4}$ & $0.7\times 10^{-4}$ & $0.5\times 10^{-4}$ & $0.6\times 10^{-4}$ & $0.2\times 10^{-4}$ & $2.0\times 10^{-4}$ & $6.0\times 10^{-3}$ & $9.0\times 10^{-3}$ & $1.8\times 10^{-3}$ & $3.5\times 10^{-2}$ & $2.1\times 10^{-3}$ & $5.9\times 10^{-3}$ & $9.0\times 10^{-3}$ & $1.8\times 10^{-2}$ & $2.9\times 10^{-2}$ & $4.3\times 10^{-2}$ & $4.0\times 10^{-2}$ & $4.4\times 10^{-2}$ \\
\Xhline{4\arrayrulewidth}
\multirow{2}*{RMSE (Trajectory Error)}
& $x(m)$ & $y(m)$ & $z(m)$ & $\alpha (rad)$ & $\beta(rad)$ & $\gamma(rad)$ & $x(m)$ & $y(m)$ & $z(m)$ & $\alpha(rad)$ & $\beta(rad)$ & $\gamma(rad)$ & $x(m)$ & $y(m)$ & $z(m)$ & $\alpha(rad)$ & $\beta(rad)$ & $\gamma(rad)$  \\
\cline{2-19}
 & $2.6\times 10^{-3}$ & $1.3\times 10^{-3}$ & $2.9\times 10^{-3}$ 
 & $1.9\times 10^{-3}$ & $1.5\times 10^{-3}$ & $2.0\times 10^{-3}$ 
 & $1.7\times 10^{-2}$ & $3.5\times 10^{-2}$ & $5.3\times 10^{-3}$ 
 & $2.9\times 10^{-3}$ & $4.1\times 10^{-2}$ & $2.9\times 10^{-3}$ 
 & $1.1\times 10^{-2}$ & $5.3\times 10^{-2}$ & $5.7\times 10^{-2}$ 
 & $1.9\times 10^{-2}$ & $1.0\times 10^{-2}$ & $4.4\times 10^{-2}$ \\
\hline
\end{tabular}
}
\end{table*}

Table~\ref{tab:controller_comparison_reduced} quantitatively demonstrates the superior performance of the MBSMC compared to the NMBSMC and PID controllers across all evaluated metrics. In terms of joint motion accuracy, MBSMC achieved RMSE values between $0.7 \times 10^{-4}~rad$ and $6.0 \times 10^{-4}~rad$, whereas NMBSMC reached up to $3.4 \times 10^{-3}~rad$ and PID reached as high as $5.2 \times 10^{-1}~rad$ over $10^{3}$ times higher than MBSMC in some joints. Velocity continuity and acceleration profiles further highlighted MBSMC’s advantage, with maximum values of $3.0 \times 10^{-3}~rad/s$ and $5.3 \times 10^{-3}~rad/s^2$, respectively, compared to NMBSMC’s $3.2 \times 10^{-2}~rad/s$ and $2.9 \times 10^{-2}~rad/s^2$, and PID’s $5.1 \times 10^{-1}~rad/s$ and $1.4 \times 10^{-1}~rad/s^2$. The jerk and snap profiles showed even more drastic differences: MBSMC’s maximum snap is $2.8 \times 10^{-4}~rad/s^4$, while NMBSMC and PID reached $3.2 \times 10^{-2}~rad/s^4$ and $2.6 \times 10^{-1}~rad/s^4$, respectively up to $10^{3}$ times higher. Steady-state error remained minimal for MBSMC (maximum $2.0 \times 10^{-4}~rad$), whereas NMBSMC and PID showed errors up to $3.5 \times 10^{-2}~rad$ and $4.4 \times 10^{-2}~rad$. Finally, in Cartesian trajectory tracking RMSE, MBSMC maintained values below $2.9 \times 10^{-3}~m$ across all axes, while NMBSMC and PID reached up to $5.3 \times 10^{-3}~m$ and $5.7 \times 10^{-2}~m$, respectively. 
To establish meaningful interpretation of the reported metrics, performance thresholds were defined in accordance with the operational demands of autonomous manipulation in chemical laboratory environments. 
Joint motion RMSE values below \(5 \times 10^{-3}\,\mathrm{rad}\) are considered sufficient for precise alignment during grasping tasks. Velocity and acceleration limits under \(1 \times 10^{-3}\,\mathrm{rad/s}\) and \(5 \times 10^{-3}\,\mathrm{rad/s^{2}}\), respectively, are necessary to ensure smooth actuation and reduce mechanical stress. For fragile object handling, jerk and snap values below \(5 \times 10^{-3}\,\mathrm{rad/s^{3}}\) and \(5 \times 10^{-3}\,\mathrm{rad/s^{4}}\) are critical to avoid abrupt force transitions. Cartesian trajectory RMSE below \(5 \times 10^{-3}\,\mathrm{m}\) is required to maintain spatial accuracy within confined workspace constraints. Based on the defined safety and performance thresholds, the presented results quantitatively demonstrate that MBSMC achieves significantly higher control precision, motion smoothness, and dynamic stability. Its performance surpasses both NMBSMC and PID by approximately three orders of magnitude and around $90$ \% less control effort in terms of motion.
\subsection{Experimental Validation}
\begin{figure}[htbp]
    \centering   \includegraphics[width=0.49\textwidth, height=2.5 in]{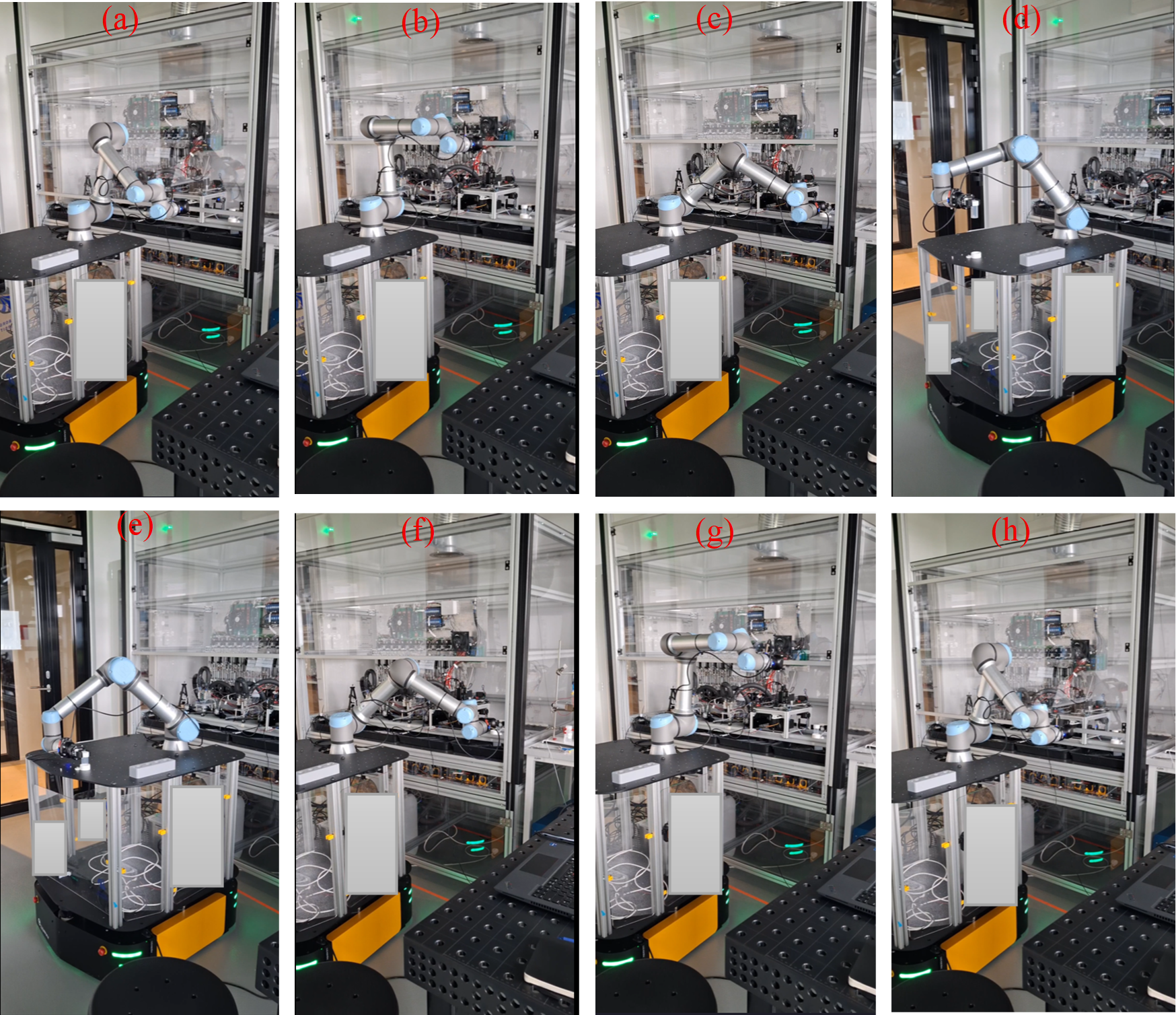}
    \caption{Motion of manipulator during experimentation controlled using MBSMC}
    \label{simn}
\end{figure}

\begin{figure}[htbp]
    \centering  \includegraphics[width=0.45\textwidth, height =1.1 in]{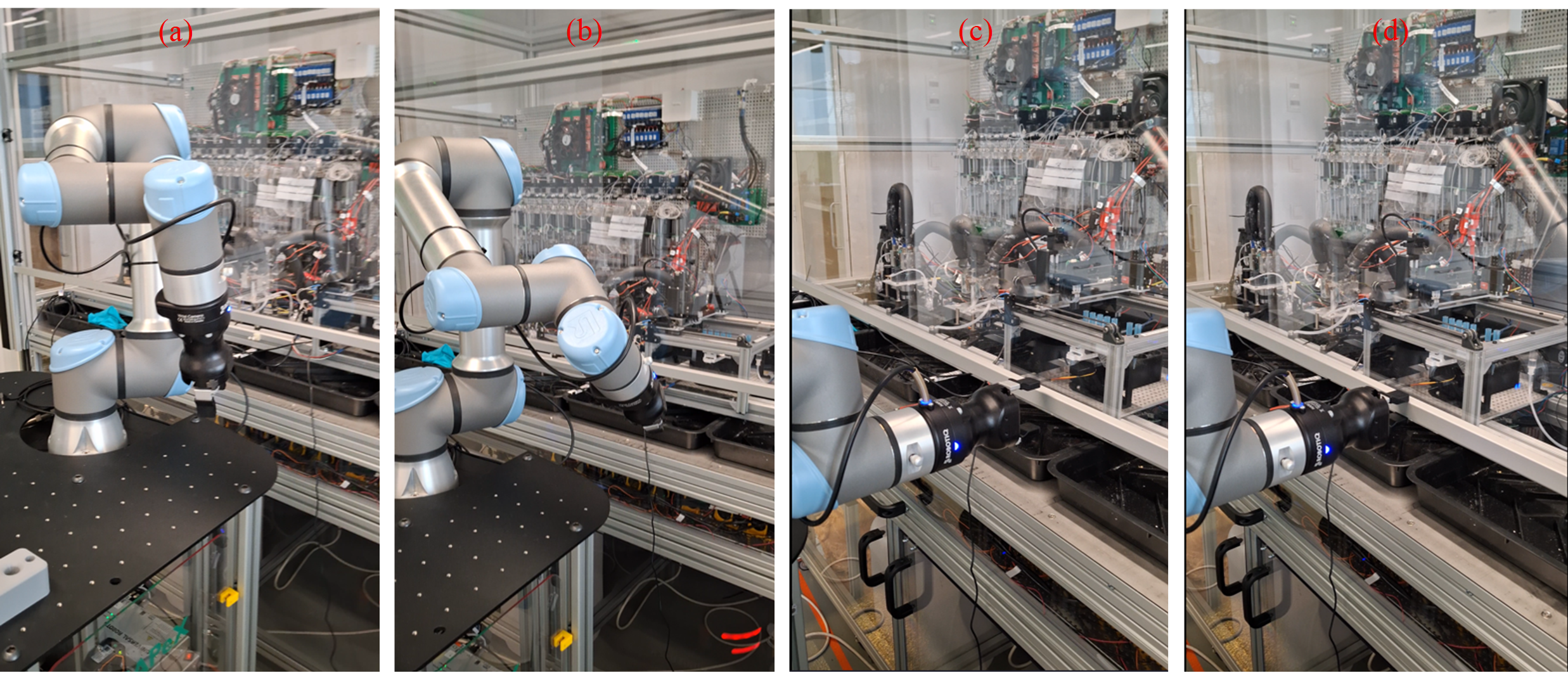}
    \caption{Opening of the window using PID controller}
    \label{sim3}
\end{figure}

\begin{figure*}[]
    \centering   \includegraphics[width=1\textwidth]{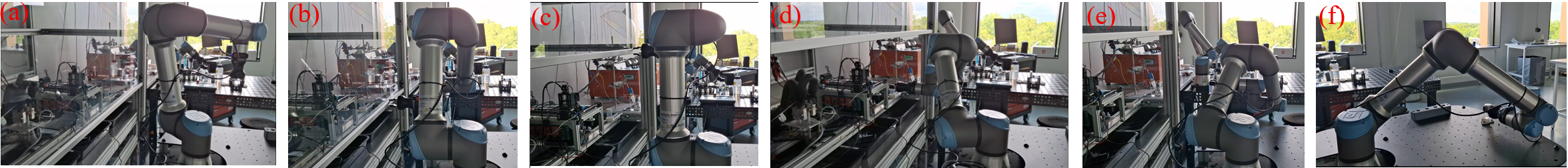}  \caption{Grasping of vessel using NMBSMC}
    \label{sim6}
\end{figure*}

\begin{figure*}[]
    \centering   \includegraphics[width=1\textwidth]{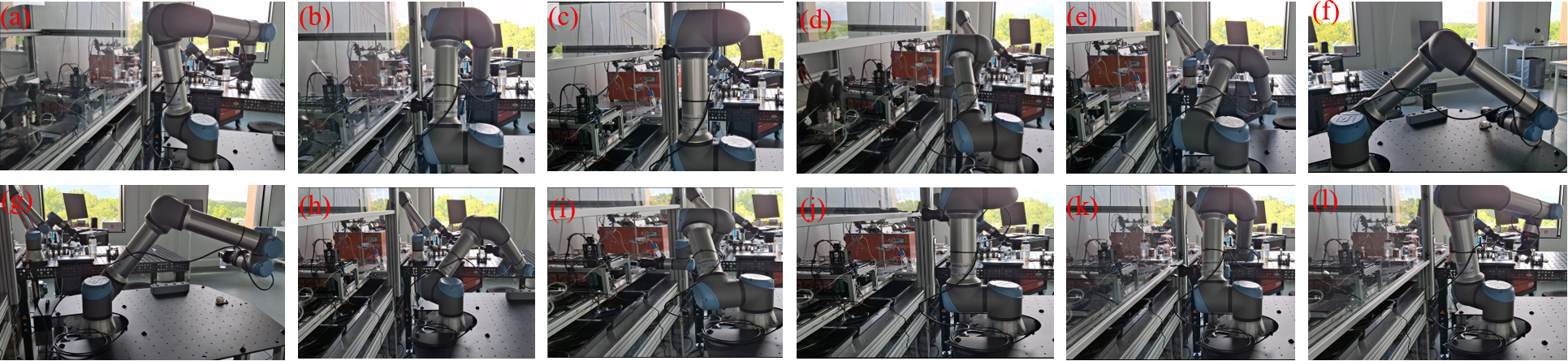}
   \caption{Motions achieved using MBSMC}
    \label{sim5}
\end{figure*}

\begin{figure}[]
    \centering   \includegraphics[width=0.35\textwidth]{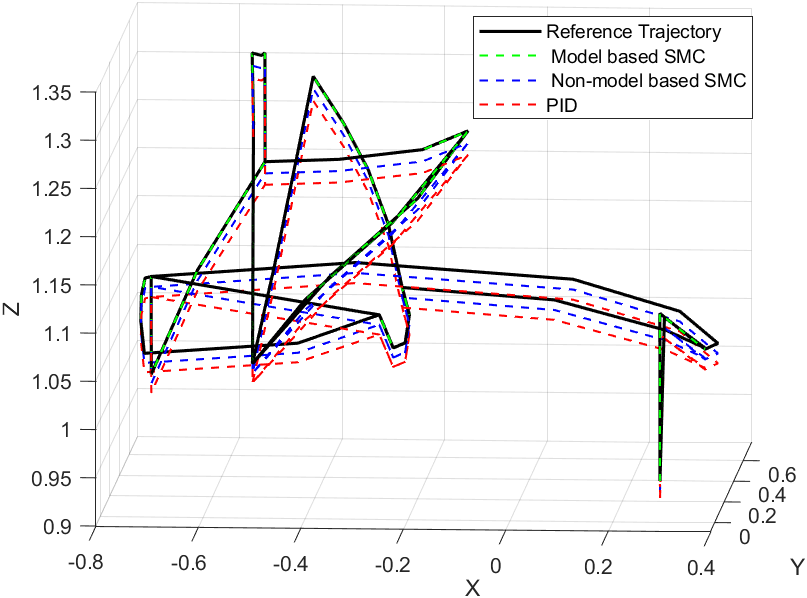}  \caption{Trajectory during experimentation using MBSMC, NMBSMC, and
PID}
    \label{sim2}
\end{figure}

\begin{figure}[]
    \centering  \includegraphics[width=0.4\textwidth]{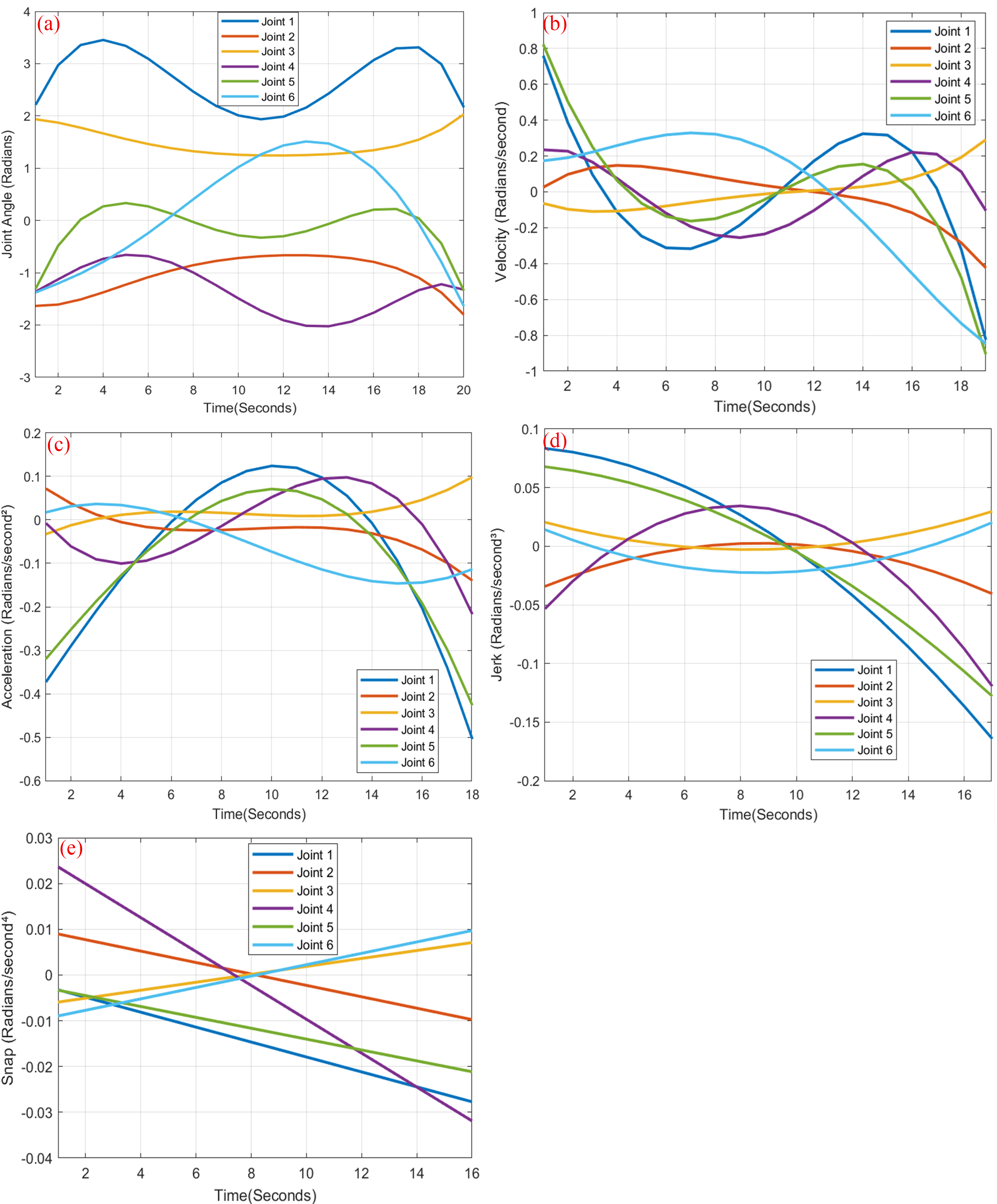}   \caption{Experimentation results using MBSMC (a)Joint angles (b)Velocity (c)Acceleration (d)Jerk (e)Snap}
    \label{sim_res1}
\end{figure}

In this study, the mobile manipulator was positioned in front of the electrolyte mixing machine, as illustrated in Fig. \ref{lab}. The mobile manipulator was tasked with picking up a vessel filled with electrolyte from the mixing system and transferring it to the ECMS machine. The process initiates once the mixing machine completes the filling operation, at which point the manipulator engages in the pick-and-place task. However, the scope of this work is limited to the control and evaluation of the manipulator’s motion under the proposed control framework. To demonstrate the controller’s effectiveness, the manipulator was instructed to perform a simplified looped motion picking the vessel and placing it back at the original location as shown in Fig. \ref{simn} enabling quantitative analysis of the controller’s performance in terms of trajectory tracking, motion stability, and precision.
The manipulator first opens the glass window (Fig. \ref{simn}(a) -(b)) and securely grasps the vessel (Fig. \ref{simn} (c)). It then places the vessel into a designated holder (Fig. \ref{simn}(e)) before returning it to the electrolyte mixing system (Fig. \ref{simn}(f)) and closes the window (Fig. \ref{simn}(g) - (h)). Notably, the grasping pose of the manipulator on the window panes is critical, as successful window operation depends heavily on the accuracy and stability of this grasp.
Both NMBSMC and MBSMC enabled appropriate grasping poses, allowing successful upward movement of the window. However, for PID-based motion control, the resulting RMSE in trajectory motion was comparatively higher, and the manipulator tended to grasp closer to the glass surface. This proximity inhibited the opening of the window and the manipulator was not able to move further as shown in Fig. \ref{sim3}. The PID controller failed to compensate for nonlinear dynamics and external disturbances, resulting in significant trajectory errors during the window-opening task, which requires precise Cartesian positioning from accurate joint angles.
\begin{figure}[htbp]
    \centering   \includegraphics[width=0.4\textwidth, height = 1.2 in]{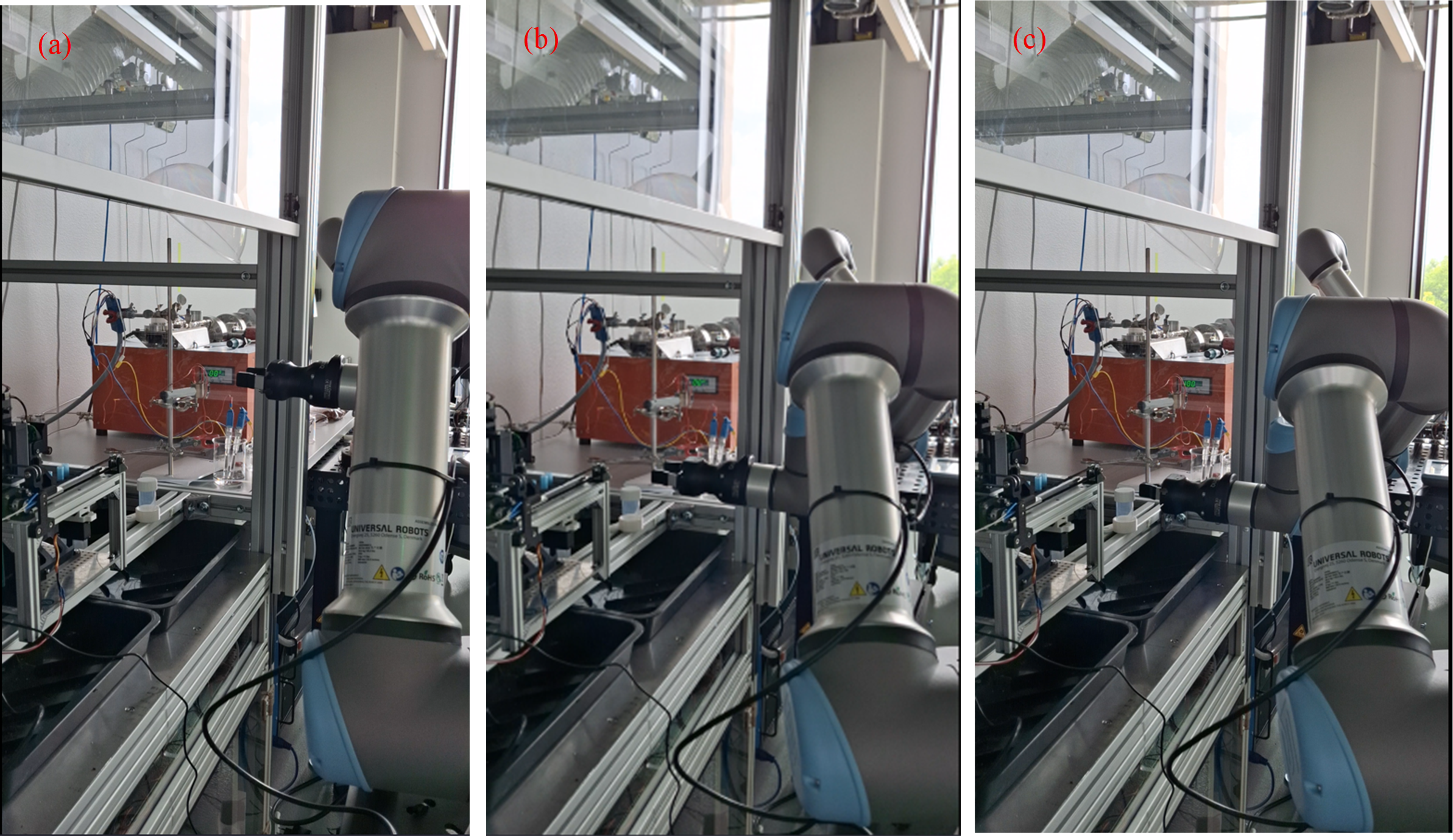}
    \caption{Grasping of the vessel using PID controller}
    \label{sim4}
\end{figure}
Similarly, as shown in Fig. \ref{sim4}, the motions governed by the PID controller failed to retrieve the vessel from the electrolyte mixing system due to RMSE in its trajectory. Although the NMBSMC successfully achieved the vessel grasping operation, it failed to accurately place the vessel onto the designated holder, as depicted in Fig. \ref{sim6}(f). 
 The NMBSMC achieved faster execution, however its poor long-term error regulation led to cumulative deviations and misalignment during vessel placement. Consequently, the vessel collided with the holder, halting the manipulator’s operation. Motion showcased using MBSMC is shown in Fig.\ref{sim5}. Compared to the other two controllers MBSMC showcased superior performance with lower RMSE and smoother cartesian motions as shown in Fig. \ref{sim2}. Comparison of trajectories traversed using 3 controllers are shown in Fig. \ref{sim_res1}. Table compares the performances of 3 controllers based on RMSE of cartesian motions, steady state error, RMSE of joint motions, VC, AP, jerk and snap values of joint motions. 

\newcolumntype{B}{!{\vrule width 1.5pt}}  
\begin{table*}[hbt!]
\centering
\caption{COMPARISON OF EXPERIMENTATION RESULTS}
\label{tab:controller_comparison}
\resizebox{0.95\textwidth}{!}{%
\begin{tabular}{|l|c|c|c|c|c|c B c|c|c|c|c|c B c|c|c|c|c|c|}
\hline
\multirow{2}{*}{\textbf{Metric}} 
 & \multicolumn{6}{c}{\textbf{MBSMC} } \setlength{\arrayrulewidth}{1.5pt}\vline
 & \multicolumn{6}{c}{\textbf{NMBSMC} } \setlength{\arrayrulewidth}{1.5pt}\vline
 & \multicolumn{6}{c}{\textbf{PID} } \setlength{\arrayrulewidth}{1.5pt}\vline \\
 \cline{2-19}
& $\theta_1$ & $\theta_2$ & $\theta_3$ & $\theta_4$ & $\theta_5$ & $\theta_6$ 
& $\theta_1$ & $\theta_2$ & $\theta_3$ & $\theta_4$ & $\theta_5$ & $\theta_6$
& $\theta_1$ & $\theta_2$ & $\theta_3$ & $\theta_4$ & $\theta_5$ & $\theta_6$ \\
\hline
RMSE of joint motions ($rad$) & $4.3\times 10^{-4}$ & $8.1\times 10^{-4}$ & $3.8\times 10^{-4}$ & $5.9\times 10^{-4}$ & $1.2\times 10^{-4}$ & $3.1\times 10^{-4}$ & $3.0\times 10^{-3}$ & $2.5\times 10^{-3}$ & $4.9\times 10^{-4}$ & $5.8\times 10^{-3}$ & $3.5\times 10^{-4}$ & $1.8\times 10^{-4}$ & $2.5\times 10^{-1}$ & $9.1\times 10^{-1}$ & $5.8\times 10^{-2}$ & $1.2\times 10^{-1}$ & $1.6\times 10^{-3}$ & $7.4\times 10^{-3}$ \\
Velocity Continuity ($rad/s$)   & $3.5\times 10^{-3}$ & $4.2\times 10^{-3}$ & $5.2\times 10^{-3}$ & $3.1\times 10^{-3}$ & $5.6\times 10^{-4}$ & $7.4\times 10^{-4}$ & $7.0\times 10^{-3}$ & $5.1\times 10^{-3}$ & $2.0\times 10^{-2}$ & $4.1\times 10^{-3}$ & $2.7\times 10^{-3}$ & $5.6\times 10^{-2}$ & $7.5\times 10^{-2}$ & $8.8\times 10^{-1}$ & $5.8\times 10^{-2}$ & $1.8\times 10^{-2}$ & $5.6\times 10^{-3}$ & $6.4\times 10^{-2}$ \\
Acceleration Profile ($rad/s^2$)  & $6.8\times 10^{-3}$ & $7.2\times 10^{-3}$ & $8.5\times 10^{-3}$ & $6.1\times 10^{-3}$ & $5.1\times 10^{-4}$ & $6.2\times 10^{-4}$ & $9.1\times 10^{-3}$ & $2.2\times 10^{-2}$ & $7.1\times 10^{-3}$ & $3.1\times 10^{-2}$ & $6.3\times 10^{-3}$ & $4.9\times 10^{-2}$ & $2.5\times 10^{-1}$ & $8.1\times 10^{-2}$ & $1.8\times 10^{-2}$ & $2.2\times 10^{-1}$ & $0.6\times 10^{-1}$ & $9.4\times 10^{-2}$ \\
Jerk Profile  ($rad/s^3$)         & $3.5\times 10^{-4}$ & $7.2\times 10^{-4}$ & $2.8\times 10^{-3}$ & $3.1\times 10^{-3}$ & $2.6\times 10^{-4}$ & $3.4\times 10^{-4}$ & $1.0\times 10^{-2}$ & $1.5\times 10^{-2}$ & $3.5\times 10^{-3}$ & $7.2\times 10^{-3}$ & $2.8\times 10^{-2}$ & $3.1\times 10^{-3}$ & $2.6\times 10^{-1}$ & $3.4\times 10^{-2}$ & $1.0\times 10^{-2}$ & $1.5\times 10^{-2}$ & $1.8\times 10^{-1}$ & $6.1\times 10^{-3}$  \\
Snap Profile    ($rad/s^4$)       & $1.5\times 10^{-4}$ & $1.2\times 10^{-4}$ & $4.8\times 10^{-4}$ & $2.1\times 10^{-4}$ & $0.9\times 10^{-4}$ & $0.6\times 10^{-4}$ & $5.0\times 10^{-3}$ & $2.3\times 10^{-2}$ & $3.8\times 10^{-3}$ & $2.0\times 10^{-2}$ & $5.5\times 10^{-2}$ & $2.8\times 10^{-3}$ & $3.5\times 10^{-2}$ & $3.1\times 10^{-1}$ & $5.8\times 10^{-2}$ & $5.2\times 10^{-2}$ & $4.6\times 10^{-1}$ & $9.4\times 10^{-2}$ \\
Steady state error  ($rad$)   & $0.5\times 10^{-4}$ & $1.2\times 10^{-4}$ & $0.8\times 10^{-4}$ & $1.1\times 10^{-4}$ & $0.3\times 10^{-4}$ & $3.4\times 10^{-4}$ & $1.0\times 10^{-2}$ & $1.5\times 10^{-2}$ & $3.0\times 10^{-3}$ & $5.9\times 10^{-2}$ & $3.5\times 10^{-3}$ & $9.8\times 10^{-3}$ & $1.5\times 10^{-2}$ & $3.1\times 10^{-2}$ & $4.8\times 10^{-2}$ & $7.2\times 10^{-2}$ & $6.6\times 10^{-2}$ & $7.4\times 10^{-2}$ \\
\Xhline{4\arrayrulewidth}
\multirow{2}*{RMSE (Trajectory Error)}
& $x(m)$ & $y(m)$ & $z(m)$ & $\alpha(rad)$ & $\beta(rad)$ & $\gamma(rad)$ & $x(m)$ & $y(m)$ & $z(m)$ & $\alpha(rad)$ & $\beta(rad)$ & $\gamma(rad)$ & $x(m)$ & $y(m)$ & $z(m)$ & $\alpha(rad)$ & $\beta(rad)$ & $\gamma(rad)$  \\
 \cline{2-19}
& $4.5\times 10^{-3}$ & $2.2\times 10^{-3}$ & $4.8\times 10^{-3}$ 
& $3.1\times 10^{-3}$ & $2.6\times 10^{-3}$ & $3.4\times 10^{-3}$ 
& $2.9\times 10^{-2}$ & $5.8\times 10^{-2}$ & $8.8\times 10^{-3}$ 
& $4.9\times 10^{-3}$ & $6.2\times 10^{-2}$ & $7.1\times 10^{-3}$ 
& $4.3\times 10^{-2}$ & $7.8\times 10^{-2}$ & $6.1\times 10^{-1}$ 
& $2.8\times 10^{-1}$ & $9.7\times 10^{-2}$ & $8.1\times 10^{-2}$ \\

\hline
\end{tabular}
}
\end{table*}
The quantitative results presented in Table~\ref{tab:controller_comparison} clearly demonstrate that the MBSMC outperforms both the NMBSMC and PID controllers across all evaluated metrics. Specifically, MBSMC achieved significantly lower RMSE values for joint motions, ranging from \(1.2 \times 10^{-4}~rad\) to \(8.1 \times 10^{-4}~rad\), which are up to three orders of magnitude smaller than those of the PID controller and substantially better than NMBSMC. In terms of motion smoothness, MBSMC exhibits superior velocity continuity, acceleration, jerk, and snap profiles, with values consistently lower than both alternatives—often by factors of $10$ to $10^3$ indicating reduced mechanical stress and enhanced control fluidity. MBSMC achieves the lowest steady-state error, with values as small as \(0.3 \times 10^{-4}~rad\), confirming its high precision and robustness. Furthermore, trajectory tracking performance, measured by RMSE in Cartesian space, shows that MBSMC maintains errors below \(5 \times 10^{-3}~m\), whereas NMBSMC and PID exhibit errors up to \(6.2 \times 10^{-2}~m\) and \(6.1 \times 10^{-1}~m\), respectively. These results collectively validate that MBSMC delivers superior control performance in terms of accuracy, smoothness, and reliability during real-time implementations.


\section{Conclusion}
This paper has presented the development of a mobile manipulator for a self-driving chemistry laboratory, emphasizing the importance of robust control strategies for enhanced object manipulation. Through comparative analysis of MBSMC, NMBSMC, and PID controllers, results confirmed that the MBSMC exceled in handling complex dynamic environments. 
The MBSMC achieved joint motion with RMSE values between $0.7 \times 10^{-4}~ rad$ and $8.1 \times 10^{-4}~ rad$, steady-state errors as low as $0.3 \times 10^{-4}~ rad$, and Cartesian trajectory RMSE under $2.9 \times 10^{-3}~ m$. Compared to PID and NMBSMC, it reduced jerk and snap values by up to $10^{3}$ times, with maximum snap values of just $2.8 \times 10^{-4}~ rad/s^4$. Velocity continuity and acceleration profiles were also significantly smoother, with peak values of $3.0 \times 10^{-3}~ rad/s$ and $5.3 \times 10^{-3}~ rad/s^2$ respectively. This improvement is particularly critical in laboratory environments where fragile glassware and precise manipulations are involved. The use of a hyperbolic tangent function within the control law effectively suppressed chattering, resulting in smoother and safer manipulator motions by attaining performance metrics within predefined thresholds under MBSMC, unlike PID and NMBSMC.
Compared to PID controllers, which failed in tasks such as grasping and opening, the MBSMC exhibited superior robustness and adaptability. While the NMBSMC offered simplicity, it lacked the precision required for delicate operations. 
The current implementation relies on a static lab layout and limited task variety, potentially hindering scalability.
Future work will focus on adaptive control for dynamic settings, lab system integration, and predictive modeling to prevent operational failures.


\newpage

\end{document}